\documentclass{article}

\usepackage{arxiv}

\usepackage[utf8]{inputenc} 
\usepackage[T1]{fontenc}    
\usepackage{url}            
\usepackage{booktabs}       
\usepackage{amsfonts}       
\usepackage{nicefrac}       
\usepackage{microtype}      
\usepackage{lipsum}		
\usepackage{graphicx}
\usepackage[square,sort,comma,numbers]{natbib}
\usepackage{doi}

\usepackage{amsmath}
\usepackage{mathrsfs}

\usepackage{amsthm} 

\newcommand{\reffig}[1]{Figure \ref{#1}}
\newcommand{\reftab}[1]{Table \ref{#1}}
\newcommand{\refalg}[1]{Algorithm \ref{#1}}
\usepackage{algorithm}
\usepackage{algorithmic}

\usepackage{graphicx}
\usepackage{multirow}
\usepackage{booktabs}
\usepackage{makecell}
\usepackage{rotating}

\usepackage{caption}
\usepackage{subfigure}

\title{Latent Network Embedding via Adversarial Auto-encoders}

\date{} 					

\author{{Minglong Lei} \\
	Beijing University of Technology\\
	Beijing, 100124, China \\
	\And
	{Yong Shi} \\
	University of Chinese Academy of Sciences\\
	Beijing, 100190, China \\
	\And
	{Lingfeng Niu}\thanks{Corresponding author}  \\
	University of Chinese Academy of Sciences\\
	Beijing, 100190, China \\
}

\renewcommand{\headeright}{A preprint}
\renewcommand{\undertitle}{}
\renewcommand{\shorttitle}{}

\hypersetup{
pdftitle={LAAE},
pdfsubject={journal paper},
pdfauthor={Minglong Lei}
}

\begin{document}
\maketitle

\begin{abstract}
Graph auto-encoders have proved to be useful in network embedding task. However, current models only consider explicit structures and fail to explore the informative latent structures cohered in networks. To address this issue, we propose a latent network embedding model based on adversarial graph auto-encoders. Under this framework, the problem of discovering latent structures is formulated as inferring the latent ties from partial observations. A latent transmission matrix that describes the strengths of existing edges and latent ties is derived based on influence cascades sampled by simulating diffusion processes over networks. Besides, since the inference process may bring extra noises, we introduce an adversarial training that works as regularization to dislodge noises and improve the model robustness. Extensive experiments on link prediction and node classification tasks show that the proposed model achieves superior results compared with baseline models.
\end{abstract}


\section{Introduction}
Learning network representations via deep neural networks achieves promising results in recent years \cite{chang2015heterogeneous, wang2016structural}. Compared with shallow network embedding models (i.e., \cite{perozzi2014deepwalk, grover2016node2vec}), deep network embedding models exhibit superior representation ability and can extract high-level structure features. The learned embeddings can be used for downstream network analysis tasks such as node classification \cite{kipf2017semi, hamilton2017inductive} and link prediction \cite{wang2016structural, pan2018adversarially}.

\emph{Graph auto-encoders} are a series of deep neural networks which apply an auto-encoder paradigm \cite{vincent2008extracting, vincent2010stacked} to network embedding. Such framework benefits from its flexibility since the encoders and decoders can be replaced with various deep models (e.g., MLP \cite{wang2016structural} and LSTM \cite{yu2018learning}) to accomplish certain tasks. Generally, the encoder aims at mapping the global or local structure information to a low-dimensional space while the decoder serves as a generator that recovers information to approximate the input. Inspired by the encoder-decoder paradigm, substantial works devote efforts to developing diverse variants of encoders and decoders. For example, DNGR \cite{cao2016deep} builds a positive pointwise mutual information (PPMI) matrix and proposes using stacked denoising auto-encoders to learn node embeddings. NetRA \cite{yu2018learning} first samples node sequences by random walks and then uses LSTM as the encoder and decoder. Above examples demonstrate that the graph auto-encoders can be designed to handle either global structures (e.g., global adjacent matrix) or local structures (e.g., short random walk sequences).

Although current graph auto-encoder models can extract useful structural patterns in networks, they assume that networks are static and all network data can be readily fed into the learning models. However, the assumption is often violated since the real network structures are not always observable and are required to be discovered \cite{myers2010convexity,linderman2014discovering}. Specifically, there exist \emph{latent ties} in networks that capture the uncertainty or possible node dependencies which have not been observed in current edges \cite{haythornthwaite2002strong}. The latent ties can be described as the joint likelihood of transmission probabilities and transmission times \cite{shi2019diffusion}. For example, consider the epidemic spreading in a network, if an infectious source infects a node multi-hops away, there is supposed to have a latent tie between those two nodes and the strength between them describes the probability of infections and also how long it would take before the infection happens. We notice that most graph auto-encoder models overlook the latent ties or simply approximate them by transmission probabilities within random walk paths \cite{cao2016deep}, which fails to explore the rich information in latent networks.

To address the shortcomings, we present in this paper a latent auto-encoder with adversarial regularization (LAAE), which is designed as a general framework for complex network structures with sparse observations.
Our core assumption here is that network structures should also be learnable. In addition to the general \emph{feature leaning} phase accomplished by auto-encoders, we pursue an extra \emph{structure learning} phase to explore the latent network structures. Instead of approximating the latent ties by linear random walk sequences, the proposed model uses cascades to capture these complex structures. The cascades are related to the diffusion process over networks and can capture both transmission probabilities and duration times during the propagation of information. We seek to model the uncertainty and durations in latent networks and resort to network inference based on above cascades to endow latent ties and existing edges with exact strengths. Concretely, we attempt to build a \emph{latent transmission matrix} that captures the weighted dependencies between nodes. Based on the latent transmission matrix, it is natural to define a latent global unit and a latent local unit to preserve the latent information. The global unit is to directly recover the connections from a global view, while the local unit is to make sure the nodes connected via latent ties and existing edges yield similar representations in the feature space.

Besides, although the auto-encoder paradigm can learn salient patterns from network data, the auto-encoders are generally regularized to preserve specific properties \cite{wang2016structural,yu2018learning}. We notice that forcing the network structures to be learnable brings extra noises and instability, which is unfavorable for the quality of embeddings. To this end, it is desirable to further regularize the feature learning phase to improve the model robustness and flexibility \cite{dai2018adversarial}. To embed the latent structures, we use an adversarial auto-encoder \cite{makhzani2015adversarial} to learn the node representations. The adversarial training introduces generative idea into the embedding step while avoiding explicitly defining a prior distribution. The generator is trained to approximate the distribution in the embedding space of the auto-encoder. Consequently, the joint optimization of the adversarial auto-encoder can improve the robustness of leaned features and filter noises from the structure learning phase.

To summarize, the contributions of this paper is as follows,
\begin{itemize}
  \item We propose a latent network embedding method based on adversarial auto-encoders which can learn robust embeddings from informative latent structures.
  \item To discover latent structures, we propose a new strategy to learn the strengths of latent ties based on diffusion cascades which capture both transmission probabilities and times.
  \item We jointly learn from latent global information and latent local information and develop an adversarial term to avoid the influences of extra noises in the latent transmission matrix.
  \item We evaluate the proposed model in several graph analysis tasks. The results demonstrate the effectiveness of our model.
\end{itemize}

The reminder of this paper is organized as follows: Section 2 reviews the line of graph auto-encoders and also the network inference techniques related to our model. Section 3 gives the model implementation from a top-down fashion and introduces the optimization process of the model. The numerical results are reported in Section 4. We conclude our paper in Section 5.

\section{Related Work}
\subsection{Graph Auto-encoders}
The graph auto-encoders are considered as a basic framework for network analysis, especially the network embedding or network representation learning. The success of graph auto-encoders mainly lies in two aspects. First, they can be designed to assemble information from different perspectives (e.g., structures and attributes \cite{jin2019network,yu2020rich}, global view and local view \cite{wang2016structural,zhang2019depth}) by imposing certain regularization terms. Second, they are flexible to incorporate various forms of encoder and decoder models that consist of multiple levels of non-linearity. Intuitively, if the reconstructions generated from the embedding space reveal desired information (e.g., the distribution of node degrees) in networks, the encoders are supposed to be reasonable feature extractors. With the encoder-decoder scheme in mind, early shallow models based on matrix factorization \cite{belkin2002laplacian, cao2015grarep} and random walks \cite{perozzi2014deepwalk, grover2016node2vec} can also be understood within this framework \cite{hamilton2017representation}. Unsurprisingly, the encoders in shallow models are usually formulated as look-up functions which have limited representation ability.

The extensions of auto-encoder framework to deep neural networks improve the model capacity and boost the performance in network embedding. A natural choice for encoders and decoders is the multi-layer perceptions \cite{wang2016structural,cao2016deep}. For example, SDNE \cite{wang2016structural} combines the deep auto-encoder with a Laplacian regularizer and jointly captures the first-order structures and second-order structures. Since local node sequences are also critical for representing network structures, advanced models mainly focus on modeling the contextural dependencies between nodes which reveal local proximities. In general, these models use sequences-to-sequence auto-encoders that can model the node sequences sampled by random walks, where the LSTMs are used as encoders and decoders \cite{yu2018learning}. When considering attributes associated with nodes in networks, the graph convolutional networks (GCNs) are adopted as feature extractors to learn from both structural information and attribute information \cite{pan2018adversarially,wu2019feature,park2019symmetric}.

There are also considerable graph auto-encoders that fall into the category of generative models, such as variational auto-encoders (VAEs) \cite{kingma2013auto} and adversarial auto-encoders (AAEs) \cite{makhzani2015adversarial}. The variational auto-encoders attempt to embed the nodes to random vectors with a prior distribution (e.g., Gaussian) in the latent space \cite{kipf2016variational, zhang2019d, sarkar2020graph}. The parameters are learned by optimizing the evidence lower bound \cite{kipf2016variational}. In contrast, the AAEs are built under a general auto-encoder framework but with an adversarial regularization phase \cite{dai2018adversarial,pan2019learning}. For example, ANE \cite{dai2018adversarial} seeks to learn robust node features by approximating the posterior distribution defined by encoders to a given prior distribution. Our model is also built under the adversarial regularized framework but is designed for latent networks.

\subsection{Network Inference}
Network inference problem aims at inferring network edges from the observations of cascades which describe certain diffusion behaviors over graphs \cite{pouget2015inferring}, e.g., information propagation \cite{bourigault2016representation} and epidemic spreading \cite{hoffmann2019learning}. The studies of cascades relate to the influence maximization problem \cite{li2018influence} and also network evolution problem \cite{leskovec2010kronecker}. Most network inference methods are built under the time-aware assumption where the cascades are truncated by a time constraint that terminates the diffusion process \cite{farajtabar2017coevolve}. Estimating network structures boils down to discovering presence and transmission rates of edges via time-stamps and can be further built as an optimization problem by maximum likelihood estimation \cite{pouget2015inferring}. The transmission rates that adhere to edges can be denoted as conditional probabilities of infection from infected nodes to uninfected nodes \cite{myers2010convexity}. Under the homogeneous setting, the information propagates in a statistically same manner along
all edges \cite{dong2019learning}. NetInf \cite{gomez2010inferring} considers cascades as directed trees and proposes to infer the conditional transmission rates between nodes. To facilitate the optimization process, NetInf only considers the trees that are more likely to propagate in graphs. On the contrary, heterogeneous models relax the assumption and endow different transmission rates to edges. NetRate \cite{gomez2011uncovering} models the conditional likelihood between nodes as a parametric form and proposes to optimize a convex inference problem based on observed cascades.

The general network inference problem can be easily extended to various settings. For example, the Bayesian inference models assume that both exact activation times and edges are unknown and required to be inferred \cite{sefer2015convex,shaghaghian2017online}. Besides, the network inference can be integrated into other frameworks to accomplish certain tasks, such as clustering \cite{prokhorenkova2019learning} and recommendation system \cite{mukherjee2019ghostlink}.
There are also a bunch of methods focus on representation learning under information cascades, which can also be used for inferring networks \cite{li2017deepcas, wang2017community}. Unlike those methods, we simulate the information diffusion process under a given network, which can explore rich information of latent structures.

\section{Our Model}

\subsection{The Framework}

\begin{figure*}
  \centering
  \includegraphics[width=1\textwidth]{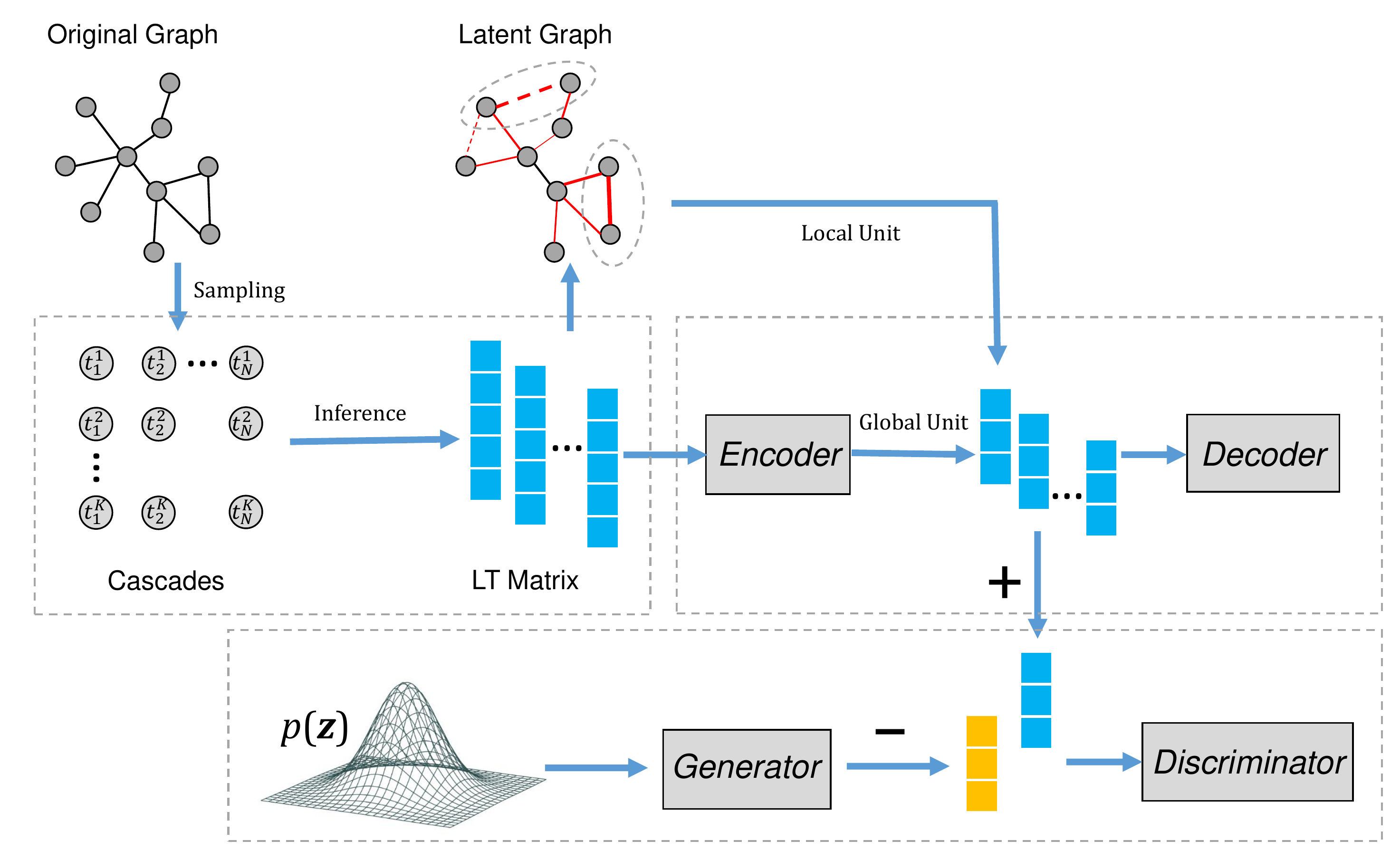}
  \caption{Overall framework of the proposed model. This framework consists of a structure learning phase (middle-left of the figure) and a feature learning phase (middle-right and bottom of the figure). The structure learning phase contains a diffusion sampling process and a network inference process. The cascades are generated to capture the high-order information. LT matrix denotes the latent transmission matrix inferred from cascades. The latent graph is built based on the LT matrix and the line widths denote the transmission rates or weights in edges. The dashed circles in the latent graph point out two types of edges. The dashed edges indicate newly inferred latent ties and the solid edges are existing edges in the original graph. The feature learning phase contains a graph auto-encoder and an adversarial regularization term. The global unit directly learns from the latent transmission matrix and keeps all connected and unconnected information via the encoder. The local unit learns from the pair-wise similarities drawn from the weighted latent graph.}
  \label{fig:framework}
\end{figure*}

We first introduce the notations and symbols that are used in this paper. Let $G(V,E)$ be a graph with a node set $V = \{v_{i}\}_{i=1}^N$ and an edge set $E \subseteq \{(v_{i},v_{j})\}_{i,j=1}^{N}$, where $N$ is the number of nodes. The adjacent matrix of graph $G$ is denoted as $A$, where $A_{i,j}=1$ represents that there is a edge between $v_{i}$ and $v_{j}$ and otherwise $A_{i,j}=0$. The Laplacian matrix is $L=D-A$ where $D\in\mathbb{R}^{N\times N}$ is a diagonal matrix $D_{i,i}=\sum_{j}A_{i,j}$. The neighbors of node $v$ is denoted as $\mathcal{N}(v)=\{u\in V | (u,v)\in E\}$. The network embedding task maps nodes in a graph to a low-dimensional space where the representation vectors preserve the node proximities.

The graph auto-encoders typically use an encoder to map the proximity information in the node domain to a hidden space, from where a decoder can recover information to approximate the input. Formally, a basic graph auto-encoder attempts to minimize the reconstruction error,
\begin{equation}\label{graph_autoencoder}
  \min \sum_{\mathbf{x}} dist(\mathbf{x},g_{\psi}(f_{\phi}(\mathbf{x}))),
\end{equation}
where $dist(\cdot)$ is a metric that measures the distance between the input and reconstruction, $f(\cdot)$ and $g(\cdot)$ are the encoder and decoder parameterized by $\phi$ and $\psi$, respectively, $\mathbf{x}$ is the input of the auto-encoder that is directly obtained from graph $G$. For example, $\mathbf{x}$ can be defined as the global similarity matrix $A$.

The proposed model is designed under the framework of graph auto-encoders but implemented for latent networks. We first present the overall framework of our model in this subsection and then give the detailed illustrations of the model components in the following subsections. Our key concerns are two-fold: First, how to discover latent structures from already existed observations; Second, how to learn robust features from latent networks while avoiding the disturbance of noises. In general, network embedding via graph auto-encoders can be formulated as the following optimization problem,
\begin{equation}\label{general_loss}
  \mathcal{L} = \mathcal{L}_{gae} + \lambda\mathcal{R},
\end{equation}
where $\mathcal{L}_{gae}$ is the  graph auto-encoder loss which learns embeddings from desired information drawn from graphs, $\mathcal{R}$ is a regularization term that preserves certain properties, e.g., robustness, $\lambda$ is a positive hyper-parameter. Under this framework, our first concern boils down to formulating inputs with latent information, which can be directly fed into the graph auto-encoders, and our second concern turns to exploring the model robustness by leveraging adversarial training as a regularization term.

An illustration of the proposed model is drawn in \reffig{fig:framework}. To address the above concerns, we combine the diffusion process with adversarial auto-encoders to solve the latent network embedding problem. Conventional network embedding methods directly learn from similarity matrices (e.g., the adjacent matrix $A$) drawn from graphs and approximate the static information. Consequently, it is desirable to unfold the undiscovered structures before feeding them into the embedding model. We refer to a structure learning phase to discover the latent structures. As shown in \reffig{fig:framework}, the diffusion sampling is used to generate cascades to capture high-order proximities in the graph. Then, a latent transmission matrix is derived by network inference from the cascades. To further learn node embeddings, we refer to a feature learning phase and use the adversarial auto-encoder to capture both global latent information and local latent information. The basic graph auto-encoder is regularized by an adversarial term which improves the model robustness.

\subsection{Structure Learning Phase}
In this subsection, we mainly discuss the structure learning phase and formulate the objective function. The goal of structure learning is to capture the latent information given the current observations. We define a weighted \emph{latent transmission matrix} $W$ where the elements represent the transmission rates between nodes. However, directly defining a learning model that infers $W$ from $A$ encounters difficulties, especially when the dimensions of $A$ and $W$ are large. To this end, we use the graph sampling method and simulate dynamic processes to capture the structural information in given static networks. We hence break the structure learning problem into two separate components. The diffusion sampling generates sequences that maximally keep the structural information while the following inference step reconstructs the connections via network inference.

Most networks embedding methods based on graph sampling use random walks to preserve local information, which are limited by their linear modeling and unstable performance. In this paper, we borrow the idea of diffusion process that has been widely used to model epidemic spreading and information propagation in networks. The diffusion process can reveal how many neighbors a node can influence in a network. The region of influence provides nice property to capture local structural information. Consequently, we can simulate diffusion processes and sample local structures from the original network. Recall that the random walks assume that a path is generated by iteratively moving from one node to another with a certain probability. This process omits the resistances of moving decided by the properties of networks. By contrast, diffusion process considers the easiness of transitions between nodes in addition to the transition probabilities. For simplicity, we follow the general diffusion setting and denote the easiness of transition as the time passing from one node to another \cite{shi2019diffusion}. At the beginning, an initially activated root node is randomly selected from the node set $V$ with time-stamp $0$. Then, the root node randomly activates a neighbor and the time interval of moving is also sampled. Then, all activated nodes in the current iteration can be used to activate their neighbors in the next iteration. After the diffusion sampling, we can obtain a set of activated nodes with time-stamps.

To facilitate the inference of latent ties, we consider all activated nodes and inactivated nodes simultaneously and use $cascades$ to denote the results of diffusion samplings. Each cascade corresponds to a diffusion sampling started at a random node. Formally, a cascade $\mathbf{c}^{i}$ is represented as an $N$-dimensional vector $(t_{1}^{i},\cdots,t_{N}^{i})$ where each element represents the activated time of a given node and $N$ is the number of nodes in a graph. Only observations within a fixed duration time are considered and all time-stamps are clipped by a time window $T$. The time-stamps for inactivated nodes are denoted as $\infty$. A set of cascades is represented as $\mathbf{C}:(\mathbf{c}^{1},\cdots,\mathbf{c}^{K}) \in \mathbb{R}^{K \times N}$ where $K$ is the number of cascades.

We can then formulate our objective function to learn the latent transmission matrix based on the sampled cascades \cite{gomez2011uncovering}. Instead of directly inferring $\phi(A;W)$, we use a bunch of observed cascades to infer the latent transmission matrix by maximizing the likelihood $\phi(\mathbf{C};W)$. We begin the discussion of the inference problem considering a single cascade $\mathbf{c} \in \mathbf{C}$ and infer its likelihood $\phi(\mathbf{c};W)$. Generally, given the time window $[0,T]$, $\phi(\mathbf{c};W)$ can be seen as a joint likelihood of observing activated nodes (i.e., $t_{i}\leq T$) and inactivated nodes (i.e., $t_{i} > T$). The pair-wise transmission from $v_{i}$ to $v_{j}$ can be modeled by an exponential distribution in the following form,
\begin{equation}
f(t_{j}|t_{i};W_{i,j})=\left\{
\begin{array}{lcl}
W_{i,j} \cdot e^{-W_{i,j}(t_{j}-t_{i})}, &  {if \ t_{i}<t_{j}}, \\
0,                                       &  {otherwise}.
\end{array}
\right.
\end{equation}
Then, the probability that $v_{j}$ is activated by $v_{i}$ can be represented as the joint probability that $v_{j}$ is activated by $v_{i}$ and is not activated by other already activated nodes in the cascade. $t_{i}$ can be interpreted as the first node to activate $t_{j}$. $\phi(t_{j}|t_{i};W)$ can be denoted as,
\begin{equation}\label{pair_likeli}
\phi(t_{j}|t_{i};W) =
   f(t_{j}|t_{i};W)
\prod_{\begin{subarray}{c} t_{k} \neq t_{i} \\ t_{k}< t_{j}\end{subarray}}
\Big{(} 1- f(t_{j}|t_{k};W) \Big{)}.
\end{equation}

From Eq. \eqref{pair_likeli}, we can calculate $\phi(t_{j};W)$ by summing up the likelihood $\phi(t_{j}|t_{i};W)$ to denote all possible activations from $v_{i}$ ($t_{i}<t_{j}$) to $v_{j}$,
\begin{equation}
\begin{split}
\phi(t_{j};W) = &
\sum_{\begin{subarray}{l}
      t_{i} < t_{j}
      \end{subarray}}
      \phi(t_{j}|t_{i};W)  \\
=&
\sum_{\begin{subarray}{l}
      t_{i} < t_{j}
      \end{subarray}}
\Big{[} f(t_{j}|t_{i};W)
\prod_{\begin{subarray}{l}
      t_{k} \neq t_{i} \\ t_{k} < t_{j}
      \end{subarray}}
\Big{(} 1-f(t_{j}|t_{k},W) \Big{)} \Big{]}.
\end{split}
\end{equation}
Then, the likelihood of observing all activations in a cascade $\phi(\mathbf{c}^{\leq T};W)$ is the joint probability of $\phi(t_{j};W)$ for all $t_{j} \leq T$,
\begin{small}
\begin{equation}
\begin{split}
\phi(\mathbf{c}^{\leq T};W)
= & \prod_{t_{j} \leq T} \phi(t_{j};W)
=   \prod_{t_{j} \leq T} \Big{(}  \sum_{t_{i} < t_{j}} \phi(t_{j}|t_{i};W) \Big{)}
\\
= & \prod_{t_j \leq T}
\Bigg{(}
\sum_{\begin{subarray}{l} t_{i} < t_{j} \end{subarray}}
  \Big{[}  f(t_{j}|t_{i};W)
     \prod_{\begin{subarray}{l} t_{k} \neq t_{i} \\ t_{k} < t_{j} \end{subarray}}
     \Big{(} 1 - f( t_{j}|t_{k};W) \Big{)}
  \Big{]}
\Bigg{)}
\\
= & \prod_{t_{j} \leq T}
\Bigg{(}
\sum_{\begin{subarray}{l} t_{i} < t_{j} \end{subarray}}
      \frac{f(t_{j}|t_{i};W)}{1- f(t_{j}|t_{i};W)}
\prod_{\begin{subarray}{l} t_{k} < t_{j} \end{subarray}}
        \Big{(} 1-f(t_{j}|t_{k};W) \Big{)}
\Bigg{)}
\end{split}
\end{equation}
\end{small}
The last step is to make the product independent of $i$ by removing the condition $k\neq i$.

The inactivated nodes $v_{l}$ ($t_{l}>T$) provide contrast information which is also useful to infer the latent structures. Consequently, the likelihood function for an entire cascade $\mathbf{c}$ can be represented as,
\begin{small}
\begin{equation}
\begin{split}
\phi(\mathbf{c};W)
= & \phi(\mathbf{c}^{\leq T};W) \times  \phi(\mathbf{c}^{> T};W)  \\
= & \prod_{t_{j} \leq T}
\Bigg{(}
\sum_{\begin{subarray}{l} t_{i} < t_{j} \end{subarray}}
\frac{f(t_{j}|t_{i};W)}{ 1-f(t_{j}|t_{i};W)}
\prod_{\begin{subarray}{l} t_{k} < t_{j} \end{subarray}}
       \Big{(} 1-f(t_{j}|t_{k};W) \Big{)}
\Bigg{)}
\\
&
\times
\Bigg{(} \prod_{\begin{subarray}{l} t_{l} > T \end{subarray}}
         \prod_{\begin{subarray}{l} t_{j} \leq T \end{subarray}}
         \Big{(}1-f(t_{l}|t_{j};W) \Big{)}
\Bigg{)},
\end{split}
\end{equation}
\end{small}
where $\phi(\mathbf{c}^{>T};W)$ denotes that nodes $v_{l}$ are not activated by already activated nodes $v_{j}$ in the cascade.

Considering the independence observation of all cascades, the estimation over all observed cascades turns to obtaining $\phi(\mathbf{C};W) = \prod_{\mathbf{c} \in \mathbf{C}} \phi(\mathbf{c};W)$. The objective required to be optimized is denoted as,
\begin{equation}\label{netinf_loss}
\mathop{\max}\limits_{W \geq 0} \sum_{\mathbf{c} \in \mathbf{C}} \log \phi(\mathbf{c};W).
\end{equation}

Following the optimization process in \cite{zhang2016inferring}, we can obtain the update rule of $W_{i,j}$ as follows:
\begin{equation}\label{solve_diffusion}
W_{i,j} =
\left\{
\begin{array}{lcl}
\frac{1}{\mathbf{N}_{c}}\sum_{\mathbf{c} \in \{\mathbf{C}: t_{j} \leq T, t_{i} < t_{j} \}} \frac{1}{t_{j}-t_{i}},
& \vspace{3mm} \\
\frac{1}{\mathbf{N}_{c}}\sum_{\mathbf{c} \in \{\mathbf{C}: t_{j} > T, t_{i} \leq T \}} \frac{1}{T-t_{i}}. \\
\end{array} \right.
\end{equation}
where $\mathbf{N}_{c}$ is a normalizer that calculates the total number of node pairs that satisfy the constraint. By solving \eqref{solve_diffusion}, we can obtain the latent transmission matrix which can be used for further feature learning.

\noindent \textbf{Information Flow Perspective}. The structure learning phase can be simply stated as building a dynamic process which starts at $A$ and ends up with an equilibrium $W$. Initially, the system is balanced and static, when we add new information into the system, i.e., starting a diffusion process at a random node, the information tends to flow through the edges and finally creates another balance system over the graph. The traces generated during the information flow record the transmission times and detect possible connections. Notice that the flow of information is a region-based spreading instead of a path-based spreading (e.g., random walk paths), our model is consequently able to capture more complex properties in the graph.

\noindent \textbf{Comparison with PPMI Matrix}. It is easy to find an analogy of the structure learning in random walk based methods which generate a probability transmission matrix named PPMI matrix \cite{cao2016deep,dai2018adversarial,qiu2018network}. PPMI matrix is based on statistic features obtained from random walks, which can be denoted as \cite{bullinaria2007extracting},
\begin{equation}
  {\rm PPMI}_{u,v}=max({\rm PMI}_{u,v},0)
\end{equation}
where ${\rm PMI}_{u,v}=\log \frac{p(u,v)}{p(u)\cdot p(v)}$ and $p(\cdot)$ is the probability. PPMI matrix depicts the statistics of co-occurrences between nodes. In contrast, the structure learning in our model attempts to develop a data-driven method to model the co-occurrences of nodes and parameterizes connections so that they can be derived from a learning process. The latent transmission matrix defines a diffusion process under a network and infers the transmission rates based on an optimization problem via maximum likelihood estimation.

\subsection{Feature Learning Phase}
The structure learning phase rebuilds the connections via inference. The following goal is to learn the embeddings that capture the proximity information distributed in the latent transmission matrix. Following the previous work \cite{wang2016structural}, we optimize a feature learning model within a global graph auto-encoder framework constrained by a local latent unit.

The global latent structure describes the common neighbors revealed by the shared nodes between $\mathcal{N}_{u}$ and $\mathcal{N}_{v}$, which is accomplished by the graph auto-encoder framework proposed in Eq. \eqref{graph_autoencoder}. The objective of global latent unit is,
\begin{equation}\label{ae}
  \mathcal{L}_{\textsc{Glo}} = \sum_{\mathbf{x}} \| \mathbf{x} - g_{\psi}(f_{\phi}(\mathbf{x})) \|_{2}^{2}.
\end{equation}
where $\mathbf{x}$ and $g_{\psi}(f_{\phi}(\mathbf{x}))$ are the input and reconstruction of the auto-encoder. We set $X=W$ to learn the global latent information from the latent transmission matrix. The embeddings are calculated by the encoder $\mathbf{y}=f_{\phi}(\mathbf{x})$. The encoder is a feature extractor which creates a distribution in the embedding space to describe the structural information in the node domain.

The local latent structure describes the direct neighbors connected by latent ties and already existing edges. Compared with previous local constrains, the key enhancement of our model is that we encourage the model to learn closer representations in the embedding space for nodes and their high-order neighbors. Besides, the local latent unit is to measure how much closeness between two nodes instead of measuring whether two nodes are close. The goal is accomplished by imposing continuous weights rather than binary codes to the local constraint. In detail, we use a local operator derived from Laplacian Eigenmaps as the local latent unit \cite{belkin2002laplacian},
\begin{equation}\label{le}
  \mathcal{L}_{\textsc{Loc}} = \sum_{1 \leq i< j \leq N} W_{i,j} \| f_{\phi}(\mathbf{x}_{i}) -f_{\phi}(\mathbf{x}_{j}) \|_{2}^{2},
\end{equation}
where $W_{i,j}$ is the weight between two nodes.

Notice that the new equilibrium obtained by structure learning brings useful latent information but also noises. Without extra regularization terms, the graph auto-encoders are unlikely to separate useful information from the noises \cite{vincent2010stacked}. To solve this problem, we propose to use an adversarial regularization to guide the auto-encoder to learn robust features \cite{makhzani2015adversarial}.

To start with, we introduce some basic concepts of adversarial methods. Training a Generative Adversarial Network (GAN) \cite{goodfellow2014generative} establishes an adversarial process based on two components, a generator $\textsc{G}_{\theta}(\cdot)$ and a discriminator $\textsc{D}_{\delta}(\cdot)$. Formally, GAN wish to optimize a min-max objective w.r.t. the generator and discriminator,
\begin{equation}\label{gan}
  \min_{\theta} \max_{\delta} \mathbb{E}_{\mathbf{x} \sim p_{d}(\mathbf{x})}[\log \textsc{D}_{\delta}(\mathbf{x})] + \mathbb{E}_{\mathbf{z} \sim p_{g}(\mathbf{z})}[\log (1-\textsc{D}_{\delta}(\textsc{G}_{\theta}(\mathbf{z})))]
\end{equation}
where $p_{d}(\mathbf{x})$ is the data distribution and $p_{g}(\mathbf{z})$ is a distribution where fake samples drawn from, $\delta$ is the parameter of discriminator and $\theta$ is the parameter of generator. The discriminator $\textsc{D}_{\delta}(\mathbf{x})$ depicts the probability that $\mathbf{x}$ comes from the real data distribution instead of the noise. Later, the Wasserstein GAN \cite{arjovsky2017wasserstein} is proposed to overcome the unstable issues of training GAN, which leverages an Earth-Mover metric to measure the distance between two distributions. The objective is then revised as,
\begin{equation}\label{wgan}
  \min_{\theta} \max_{\delta \in \Delta} \mathbb{E}_{\mathbf{x} \sim p_{d}(\mathbf{x})}[\textsc{D}_{\delta}(\mathbf{x})] - \mathbb{E}_{\mathbf{z} \sim p_{g}(\mathbf{z})}[ \textsc{D}_{\delta}(\textsc{G}_{\theta}(\mathbf{z}))],
\end{equation}
where $\Delta$ is the Lipschitz constraint that is kept by clipping the parameters within $[-c,c]$.

As illustrated in \reffig{fig:framework}, the goal of LAAE is to minimize the distance of the embedding distribution from the encoder $f_{\phi}(\mathbf{x})\sim p_{\phi}(\mathbf{x})$ and the representation distribution from the generator $G_{\theta}(\mathbf{z})\sim p_{\theta}(\mathbf{z})$. We use the Earth Moving distance $\mathcal{W}(p_{\phi}(\mathbf{x}),p_{\theta}(\mathbf{z}))$ following WGAN to measure the distance. Convert the Earth Moving distance to its dual form with smooth constraint ($\{\textsc{D}_{\delta}(\cdot)\}_{\delta \in \Delta}$ are all $K$-Lipschitz for some $K$), we can obtain,
\begin{equation}\label{wgan2}
  \mathcal{W}(p_{\phi}(\mathbf{x}),p_{\theta}(\mathbf{z})) \approx \max_{\delta \in \Delta} \mathbb{E}_{\mathbf{x} \sim p_{d}(\mathbf{x})}[\textsc{D}_{\delta}(f_{\phi}(\mathbf{x}))] - \mathbb{E}_{\mathbf{z} \sim p_{g}(\mathbf{z})}[\textsc{D}_{\delta}(\textsc{G}_{\theta}(\mathbf{z}))].
\end{equation}

Under the min-max optimization framework, the discriminator and generator can be trained separately. The discriminator loss function can be denoted as,
\begin{equation}\label{dis}
\mathcal{L}_{\textsc{D}}(\delta) = -\mathbb{E}_{\mathbf{x} \sim p_{d}(\mathbf{x})}[\textsc{D}_{\delta}(f_{\phi}(\mathbf{x}))] + \mathbb{E}_{\mathbf{z} \sim p_{g}(\mathbf{z})}[\textsc{D}_{\delta}(\textsc{G}_{\theta}(\mathbf{z}))].
\end{equation}
The generator loss function can be written as,
\begin{equation}\label{gen}
\mathcal{L}_{\textsc{G}}(\theta) = \mathbb{E}_{\mathbf{x} \sim p_{d}(\mathbf{x})}[\textsc{D}_{\delta}(f_{\phi}(\mathbf{x}))] - \mathbb{E}_{\mathbf{z} \sim p_{g}(\mathbf{z})}[\textsc{D}_{\delta}(\textsc{G}_{\theta}(\mathbf{z}))].
\end{equation}

In summary, we aim to jointly minimize the latent global loss, latent local loss and the adversarial regularization process. The objective can be denoted in the following form,
\begin{equation}\label{loss}
  \mathcal{L}(\phi,\psi,\theta,\delta) = \mathcal{L}_{\textsc{Glo}} + \lambda_{1}\mathcal{L}_{\textsc{Loc}} + \lambda_{2}\mathcal{W}(p_{\phi}(\mathbf{x}),p_{\theta}(\mathbf{z})),
\end{equation}
where $\lambda_{1}$ and $\lambda_{2}$ are two positive hyper-parameters. We can then derive the gradients of the parameters as follows,
\begin{align}
\begin{split}\label{eq:1}
\nabla_{\phi}\mathcal{L} &= \nabla_{\phi} \sum_{\mathbf{x}} \| \mathbf{x} - g_{\psi}(f_{\phi}(\mathbf{x})) \|_{2}^{2} \\
& + \lambda_{1} \nabla_{\phi} \sum_{1\leq i<j\leq N} W_{i,j} \| f_{\phi}(\mathbf{x}_{i}) -f_{\phi}(\mathbf{x}_{j}) \|_{2}^{2} \\
& + \lambda_{2} \nabla_{\phi} \mathbb{E}_{\mathbf{x} \sim p_{d}(\mathbf{x})}[\textsc{D}_{\delta}(f_{\phi}(\mathbf{x}))]
\end{split}\\
\begin{split}\label{eq:2}
\nabla_{\psi}\mathcal{L} &= \nabla_{\psi} \sum_{\mathbf{x}} \| \mathbf{x} - g_{\psi}(f_{\phi}(\mathbf{x})) \|_{2}^{2}
\end{split}\\
\begin{split}\label{eq:3}
\nabla_{\delta}\mathcal{L} &= -\lambda_{2} \nabla_{\delta} \mathbb{E}_{\mathbf{x} \sim p_{d}(\mathbf{x})}[\textsc{D}_{\delta}(f_{\phi}(\mathbf{x}))] + \lambda_{2} \nabla_{\delta} \mathbb{E}_{\mathbf{z} \sim p_{g}(\mathbf{z})}[\textsc{D}_{\delta}(\textsc{G}_{\theta}(\mathbf{z}))]
\end{split}\\
\begin{split}\label{eq:4}
\nabla_{\theta}\mathcal{L} &= -\lambda_{2} \nabla_{\delta} \mathbb{E}_{\mathbf{z} \sim p_{g}(\mathbf{z})}[\textsc{D}_{\delta}(\textsc{G}_{\theta}(\mathbf{z}))]
\end{split}
\end{align}
Based on above derivatives, we can alternatively optimizing the different components of LAAE by block coordinate descent following \cite{yu2018learning}. The training of encoder (update $\phi$) is not only determined by the locally regularized reconstruction phase that captures latent information, but also influenced by the adversarial process that improves the robustness of learned embeddings. The overall algorithm for our model is listed in \refalg{alg1}.

\floatname{algorithm}{Algorithm}
\begin{algorithm}[!h]
  \caption{Latent Adversarial Auto-Encoder Training}
  \label{alg1}
  \begin{algorithmic}[1]
    \REQUIRE Graph $G(V,E)$, adjacent matrix $A$,
             hyper-parameter $\lambda_{1}$, $\lambda_{2}$, number of iterations $iters$, batch size $B$
    \STATE Run diffusion sampling process to obtain $K$ cascades $\mathbf{C}:(\mathbf{c}^{1},\cdots,\mathbf{c}^{K})$;
    \STATE Solve Eq. \eqref{netinf_loss} to obtain the diffusion information matrix $W$;
    \STATE Set $X = W$;
    \FOR {$iter<iters$}
       \STATE \textbf{[Training Graph Auto-encoder]}
       \STATE Draw samples $\{\mathbf{x}\}_{i=1}^{B}$ from the data distribution $p_{d}(\mathbf{x})$;
       \STATE Calculate the loss of graph auto-encoder $\mathcal{L}_{gae}=\mathcal{L}_{\textsc{Glo}}+\lambda_{1}\mathcal{L}_{\textsc{Loc}}$;
       \STATE Update the parameters $\phi$ and $\psi$ for the encoder and decoder by \eqref{eq:1} and \eqref{eq:2};
       \STATE \textbf{[Training Discriminator]}
       \STATE Draw positive samples $\{\mathbf{x}\}_{i=1}^{B}$ from the data distribution $p_{d}(\mathbf{x})$;
       \STATE Draw negative samples $\{\mathbf{z}\}_{i=1}^{B}$ from a prior distribution $p_{g}(\mathbf{z})$;
       \STATE Calculate $f_{\phi}(\mathbf{x})$ and $\textsc{G}_{\theta}(\mathbf{z})$;
       \STATE Calculate the loss of discriminator $\mathcal{L}_{\textsc{D}}$ by \eqref{dis};
       \STATE Update parameters of discriminator $\delta$ by \eqref{eq:3};
       \STATE Clip the parameters $\delta$ within $[-c,c]$;
       \STATE \textbf{[Training Generator]}
       \STATE Draw negative samples $\{\mathbf{z}\}_{i=1}^{B}$ from a prior distribution $p_{g}(\mathbf{z})$;
       \STATE Calculate $\textsc{G}_{\theta}(\mathbf{z})$;
       \STATE Calculate the loss of Generator $\mathcal{L}_{\textsc{G}}$ by \eqref{gen};
       \STATE Update parameters of generator $\theta$ by \eqref{eq:4};
    \ENDFOR
  \end{algorithmic}
\end{algorithm}

\section{Experiments}
To demonstrate the effectiveness of the proposed model, we evaluate our model in link prediction and node classification tasks.

\begin{table}[htbp]
  \centering
  \caption{Dataset Statistics}
  \label{tab:dataset}
    \begin{tabular}{lcccl}
    \hline
    Datasets & $\#$ Nodes & $\#$ Edges & $\#$ Labels \\
    \hline
    PPI           & 3,890   & 76,584      & None \\
    ADV           & 5,155   & 39,285      & None \\
    JDK           & 6,434   & 53,892      & None \\
    BLOG          & 10,312  & 333,983     & None \\
    DBLP          & 2,760   & 3,818       & 4    \\
    BLOG5K        & 5,196   & 171,743     & 6    \\
    \hline
    \end{tabular}%
\end{table}%

\subsection{Experiment Setup}
We conduct the experiments under 6 different datasets. The statistics of all datasets are listed in \reftab{tab:dataset}. All networks only contain structural information. The PPI network, JDK network, ADV network and BLOG network are used for link prediction while the DBLP network and BLOG5K network are used for node classification.
\begin{itemize}
  \item Protein-Protein Interactions (PPI) \cite{grover2016node2vec,breitkreutz2007biogrid} is a biological network which describes the directed interactions between proteins. The network used in this paper is a subgraph of the PPI network for Homo Sapiens.
  \item Advogato (ADV)\footnote{\url{http://konect.uni-koblenz.de/networks/advogato}} \cite{massa2009bowling} is a trust network built under an online community platform named Advogato. The nodes are the users and the directed edges are their trust relationships.
  \item JDK Dependency (JDK)\footnote{\url{http://konect.uni-koblenz.de/networks/subelj_jdk}} \cite{yu2018learning} is a software dependency network for JDK 1.6.0.7 framework. The nodes are the classes while the edges are the directed dependencies between them.
  \item Blogcatalog (BLOG) \cite{tang2009relational} is an undirected social network that depicts the social relations between bloggers. The nodes are the bloggers and the edges indicate their friendships.
  \item DBLP \cite{sun2009ranking} is an academic network in computer science. We use the sub-network of DBLP which only contains four areas. Five representative conferences are selected from each area. The nodes are the authors that have published papers in these conferences and the edges are their co-authorships.
  \item Blogcatalog5k (BLOG5K) \cite{huang2017label} is a variant of Blogcatalog network which contains 5196 nodes associated with labels. The labels are selected from pre-defined classes and describe the interests of the bloggers.
\end{itemize}

We compare the proposed model with several baseline models that are widely used in network embedding.
\begin{itemize}
  \item DeepWalk \cite{perozzi2014deepwalk} is a local network embedding method which depends on random walk samplings. The goal is to map the proximity within short node sequences to a low-dimensional space.
  \item node2vec \cite{grover2016node2vec} is another random walk based model, which extends the sampling strategy in DeepWalk to its biased version which contains Depth-First-Samplings and Breadth-First-Samplings.
  \item Spectral Clustering (SC) \cite{ng2002spectral} or spectral embedding applies spectral decomposition to the Laplacian matrix of the network. The embeddings are accumulated by the top eigenvectors of the Laplacian matrix.
  \item Structural Deep Network Embedding (SDNE) \cite{wang2016structural} is a deep network embedding method which leverages a graph auto-encoder and a local constraint to learn the node representations, so that the local information and global information can both be captured.
  \item Adversarial Network Embedding (AAE) \cite{dai2018adversarial} is a deep graph auto-encoder framework which introduces adversarial training into the model.
  \item NetRA \cite{yu2018learning} is also an adversarially regularized graph auto-encoder based on random walk sequences. This framework uses long short-term memory network (LSTM) as an encoder to preserve local structures.
\end{itemize}

For our model, to generate sufficient cascades, all nodes in the network are set as root nodes. The maximal time window $T$ is set as $10$.

\subsection{Link Prediction}

We first apply the proposed model in the link prediction task to evaluate whether the learned representations can capture the proximities and structures in the node domain. The goal of link prediction is to predicted missing edges in a network where part of its edges have been removed. We can then consider the link prediction task as a supervised classification framework based on edge features. We randomly removed 50$\%$ edges for all networks while keeping the remaining network connected. An equal number of negative samples are also sampled from the node pairs without any edges.

Once the node representation vectors are computed after the unsupervised training, we use an operator to generate the edge features $\mathbf{e}=op(\mathbf{y}(u),\mathbf{y}(v))$ where $\mathbf{y}(u)$ and $\mathbf{y}(v)$ are the low-dimensional vectors of node $u$ and $v$, and $\mathbf{e}$ is the edge features between them. We select Hadamard and Weighted-L2 as operators following the setting in node2vec \cite{grover2016node2vec}. The Hadamard operator $\diamond$ is defined as $[\mathbf{y}(u)\diamond\mathbf{y}(v)]_{i}=\mathbf{y}_{i}(u) \ast \mathbf{y}_{i}(v)$, and the Weighted-L2 operator $\|\cdot\|_{\bar{2}}$ is defined as $\| \mathbf{y}(u) \cdot \mathbf{y}(v)\|_{\bar{2}i}=| \mathbf{y}_{i}(u) - \mathbf{y}_{i}(v) |^{2}$.

After obtaining all positive and negative edge vectors, the link prediction can be considered as a binary classification task. We draw the Receiver Operating Characteristic (ROC) curves and report the Area Under Curve (AUC) scores for all models on PPI network, ADV network, JDK network, and BLOG network.

The AUC scores are listed in \reftab{tab:auc_scores}. It can be observed that LAAE outperforms all baseline models including shallow models and recently proposed deep graph auto-encoders. LAAE achieved at least 2$\%$ performance improvements on all datasets. It is noticeable that introducing adversarial training to original networks not necessarily improves the performances in some datasets. For example, SDNE obtains higher results on BLOG compared with AAE and NetRA with Hadamard operator. In addition, the performances of different models may vary w.r.t. different operators. On PPI network, NetRA outperforms AAE with Hadamard operator and obtains opposite results with Weighted-L2 operator. The proposed LAAE obtains best performance in both operators, which illustrates that the learned embeddings capture underlying information in networks.

\begin{table}[htbp]
  \centering
  \footnotesize
  \caption{Link Prediction Results w.r.t. AUC scores.}
  \begin{tabular}{llcccc}
  \Xhline{2\arrayrulewidth}
  Operator & Methods & PPI & ADV & JDK & BLOG\\
  \Xhline{2\arrayrulewidth}
  \multirow{5}{*}{Hadamard}
  & Deepwalk       & 0.6795 & 0.7197 & 0.8401 & 0.6821 \\
  & node2vec       & 0.7259 & 0.7451 & 0.8652 & 0.7295 \\
  & Spectral       & 0.7302 & 0.7901 & 0.5952 & 0.6032 \\
  & SDNE           & 0.7782 & 0.8318 & 0.8863 & 0.8861 \\
  & AAE            & 0.8281 & 0.8527 & 0.9293 & 0.8661 \\
  & NetRA          & 0.8321 & 0.8980 & 0.9148 & 0.8532 \\
  & \textbf{LAAE}  & \textbf{0.8752} & \textbf{0.9105} & \textbf{0.9444} & \textbf{0.9473} \\
  \hline
  \multirow{5}{*}{Weighted-L2}
  & Deepwalk       & 0.7715 & 0.8414 & 0.8472 & 0.8596 \\
  & node2vec       & 0.8048 & 0.8570 & 0.8445 & 0.8897 \\
  & Spectral       & 0.7179 & 0.7493 & 0.8551 & 0.7602 \\
  & SDNE           & 0.7282 & 0.7959 & 0.8643 & 0.7235 \\
  & AAE            & 0.8295 & 0.8590 & 0.8515 & 0.7626 \\
  & NetRA          & 0.7816 & 0.8500 & 0.8947 & 0.9093 \\
  & \textbf{LAAE}  & \textbf{0.8404} & \textbf{0.8821} & \textbf{0.9417} & \textbf{0.9312} \\
  \Xhline{2\arrayrulewidth}
  \end{tabular}
  \label{tab:auc_scores}
\end{table}

\begin{figure}[htbp]
  \centering
  \includegraphics[width=1\textwidth,trim={50 0 50 0},clip]{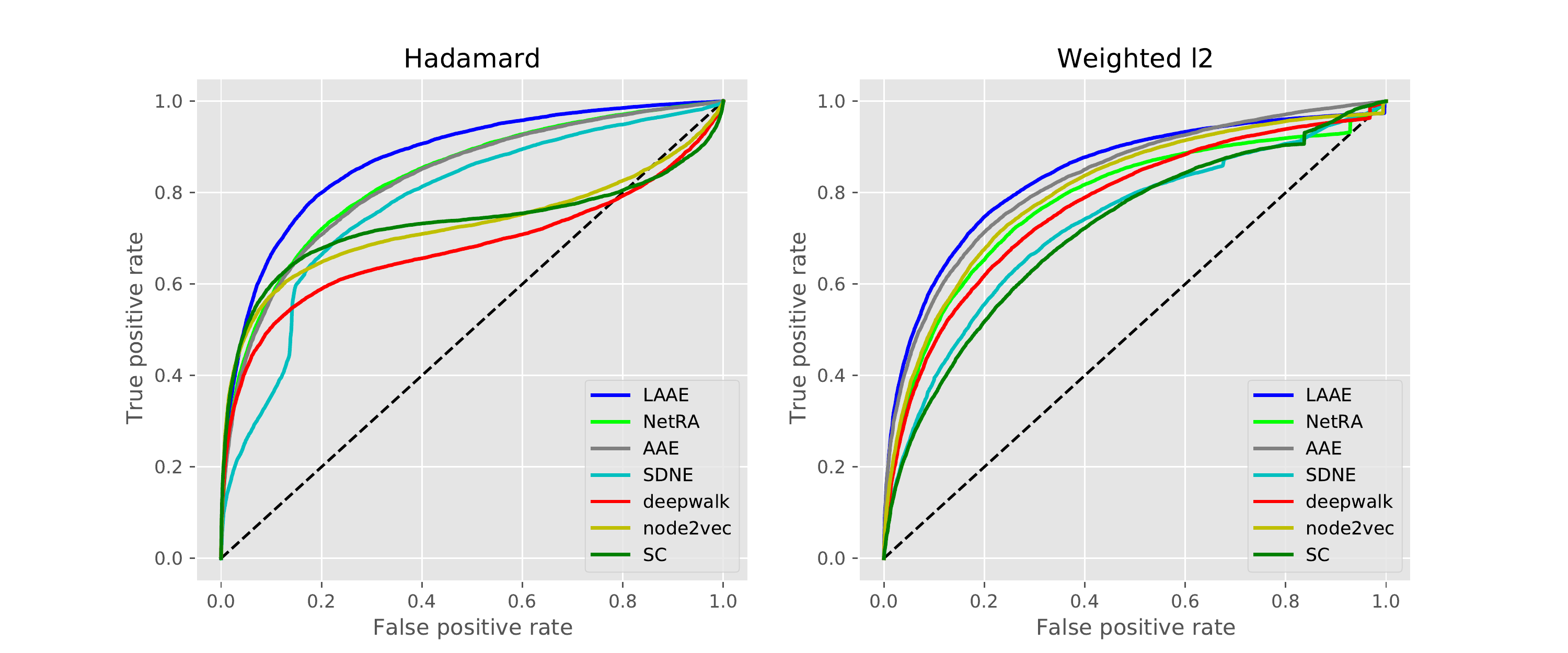}
  \caption{ROC curves on PPI network with Hadamard operator and Weighted L2 operator.}
  \label{fig:ppi_roc}
\end{figure}

\begin{figure}[htbp]
  \centering
  \includegraphics[width=1\textwidth,trim={50 0 50 0},clip]{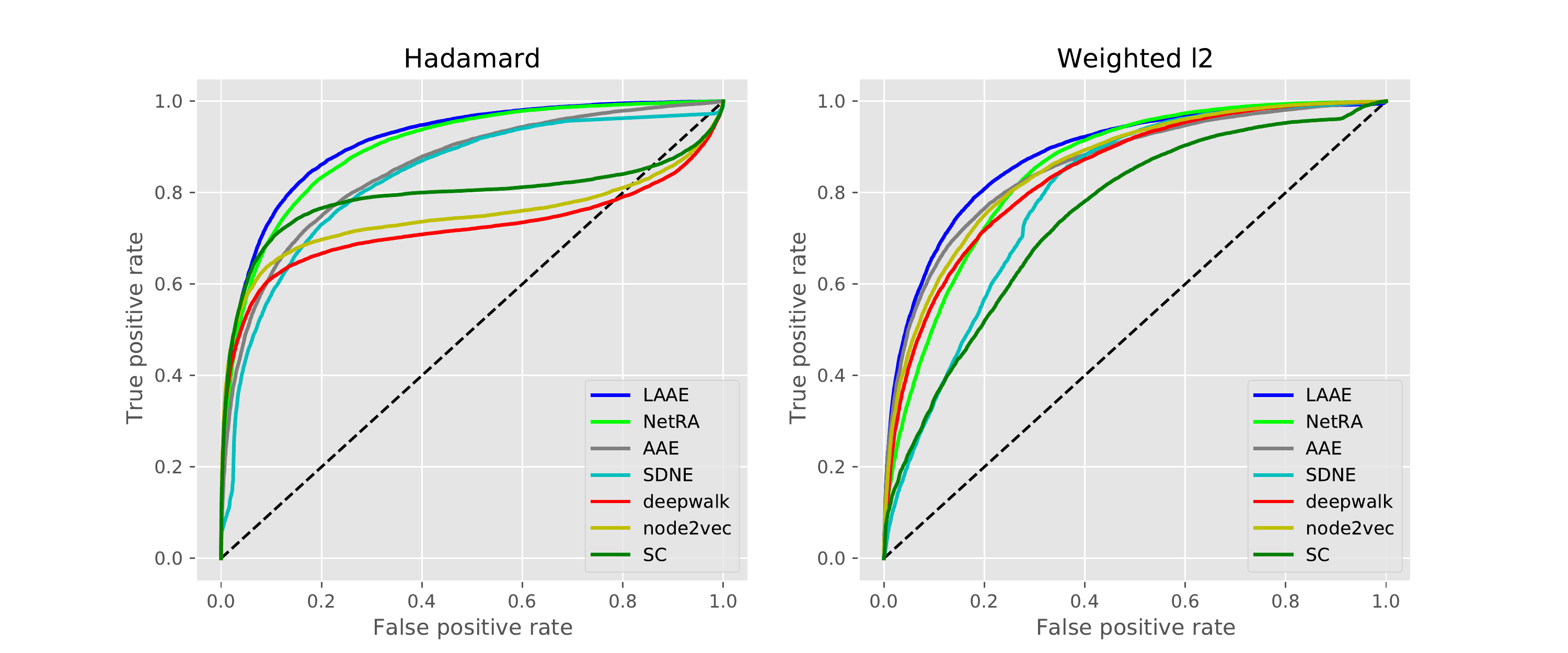}
  \caption{ROC curves on ADV network with Hadamard operator and Weighted L2 operator.}
  \label{fig:adv_roc}
\end{figure}

\begin{figure}[htbp]
  \centering
  \includegraphics[width=1\textwidth,trim={50 0 50 0},clip]{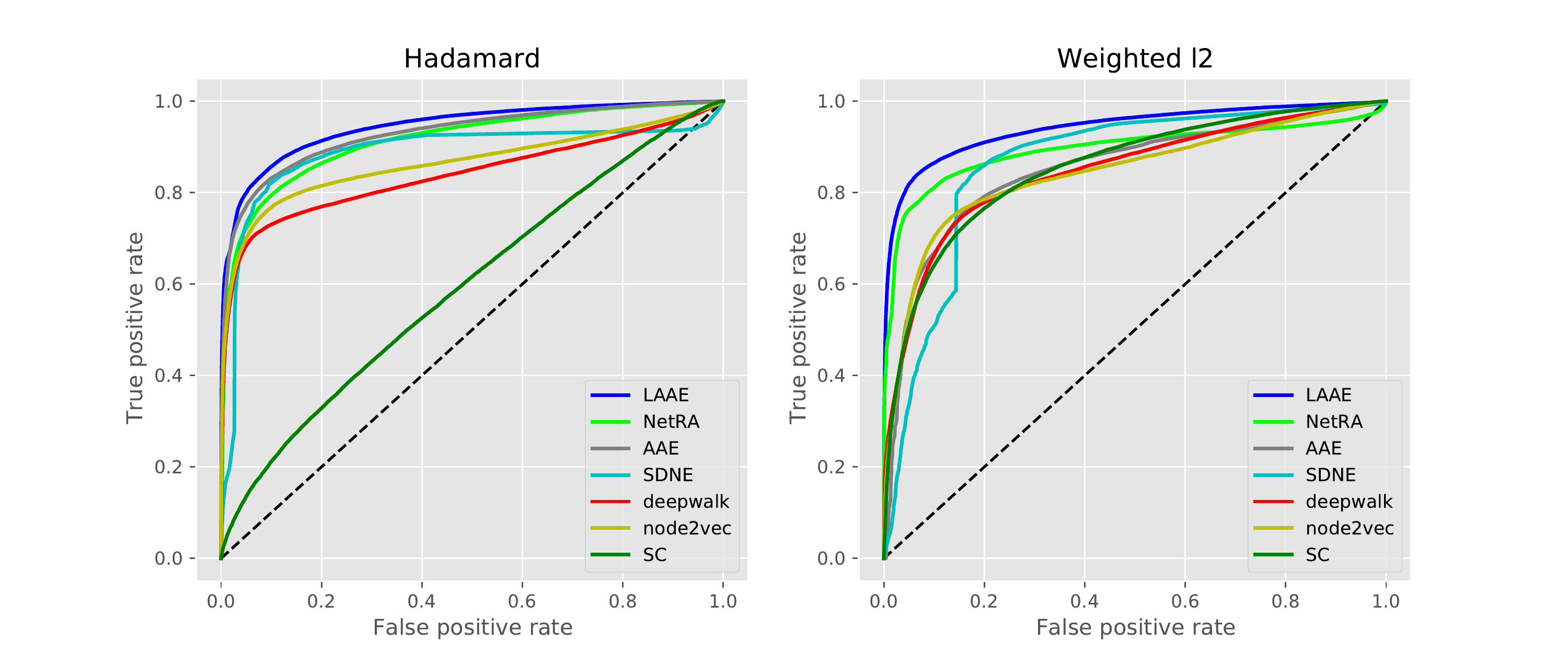}
  \caption{ROC curves on JDK network with Hadamard operator and Weighted L2 operator.}
  \label{fig:jdk_roc}
\end{figure}

\begin{figure}[htbp]
  \centering
  \includegraphics[width=1\textwidth,trim={50 0 50 0},clip]{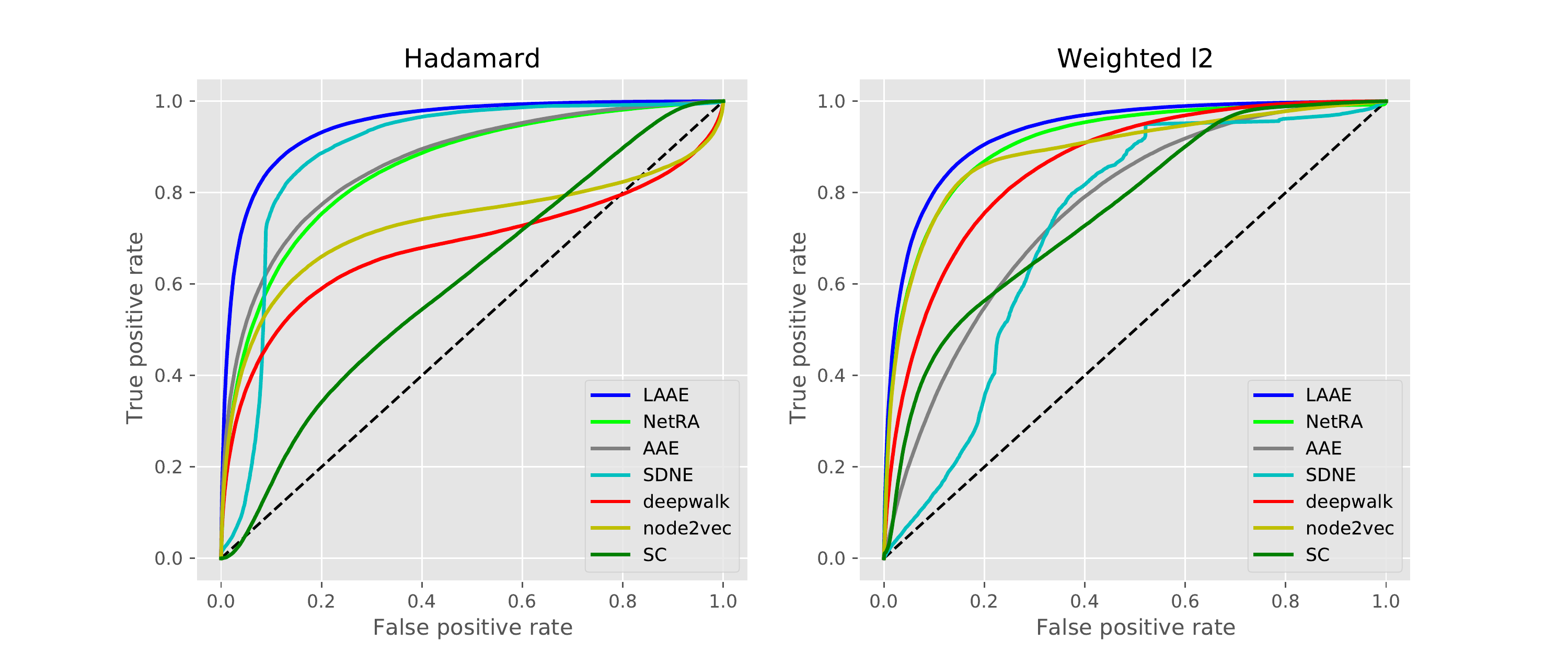}
  \caption{ROC curves on BLOG network with Hadamard operator and Weighted L2 operator.}
  \label{fig:blog_roc}
\end{figure}

The ROC curves on PPI, ADV, JDK, and BLOG are also plotted in \reffig{fig:ppi_roc}, \reffig{fig:adv_roc}, \reffig{fig:jdk_roc}, \reffig{fig:blog_roc}, respectively. The curves of the proposed LAAE are very close to the top-left corner compared with all baseline models, which indicates better representation ability for our model from another perspective.

\subsection{Node Classification}

In this subsection, we apply the representation vectors learned from our model to node classification task. Predicting the categories of nodes is a fundamental task in graph-based studies. We follow the experimental settings in DeepWalk and feed the learned features into the one-vs-rest logistic regression model \cite{fan2008liblinear}. To test the quality of features, the training samples used in the supervised phase are randomly split for training and test. The ratio of training samples ranges from 10$\%$ to 90$\%$. We report the average Micro-F1 scores and Macro-F1 scores after ten runs for our model and baseline models on DBLP network and BLOG5K network.

\begin{figure}[htbp]
  \centering
  \includegraphics[width=1\textwidth,trim={50 0 50 0},clip]{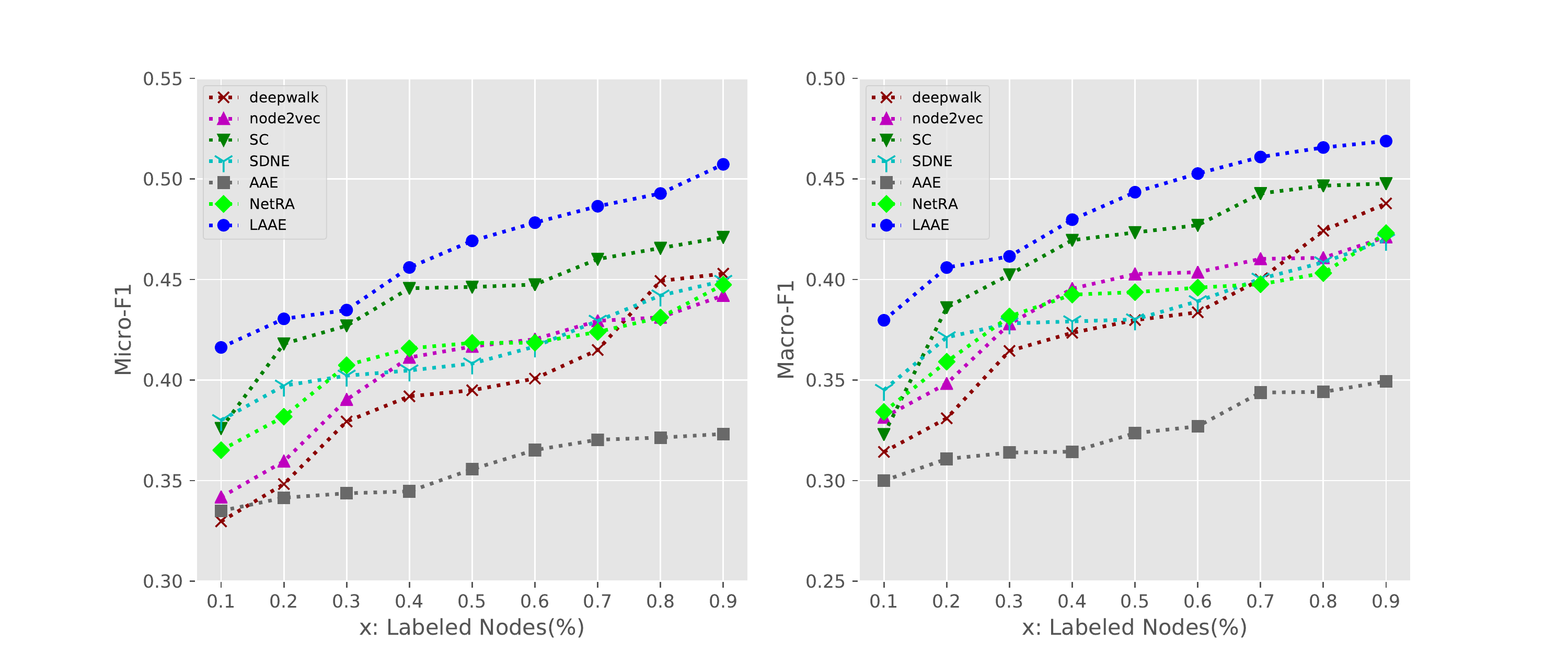}
  \caption{Classification results on DBLP w.r.t. Micro-F1 and Macro-F1 scores with the training ratio varies from $10\%$ to $90\%$.}
  \label{fig:dblp_classification}
\end{figure}

\begin{figure}[htbp]
  \centering
  \includegraphics[width=1\textwidth,trim={50 0 50 0},clip]{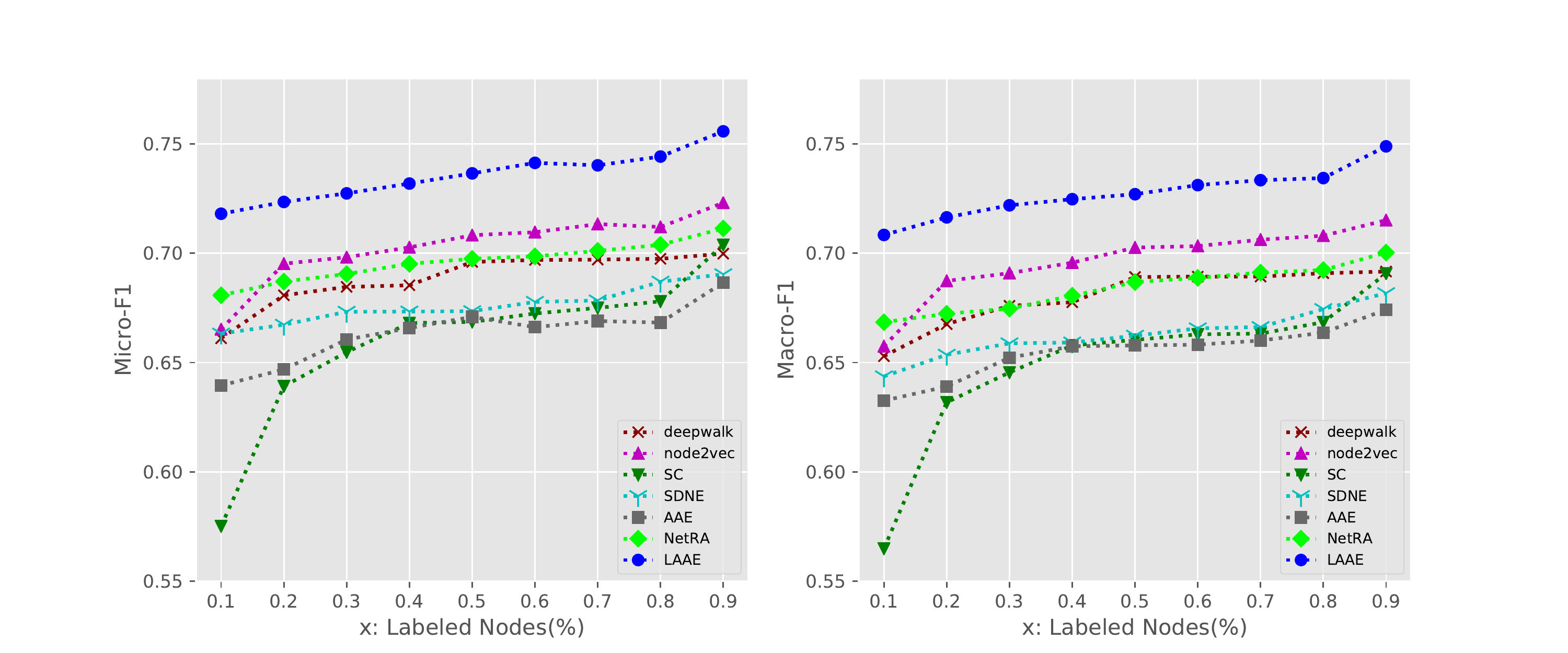}
  \caption{Classification results on BLOG5K w.r.t. Micro-F1 and Macro-F1 scores with the training ratio varies from $10\%$ to $90\%$.}
  \label{fig:blogcatalog_classification}
\end{figure}

The performances are shown in \reffig{fig:dblp_classification} and \reffig{fig:blogcatalog_classification}, respectively. LAAE achieves better classification results in both datasets, which indicates the learned embeddings capture useful information in networks. The performance gains in BLOG5K are more significant than that in DBLP. We also notice that, in a denser network such as BLOG5K, deep models achieve more stable results with respect to different ratios of training samples. LAAE is less affected by the sparsity of networks since the structure learning phase detects latent ties and increases the information in networks.

\subsection{Model Analysis and Parameter Study}

\begin{figure}[htbp]
  \centering
  \includegraphics[width=1\textwidth,trim={50 0 50 0},clip]{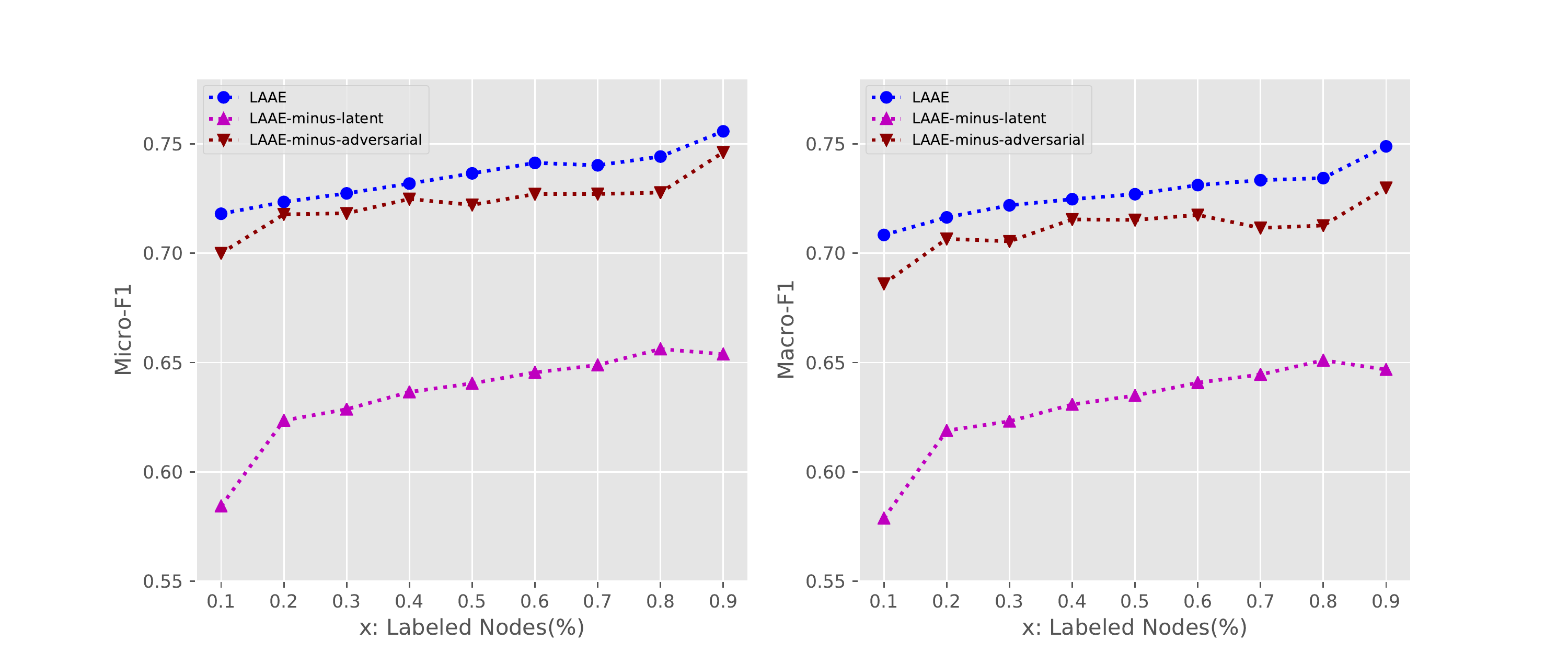}
  \caption{Comparisons between LAAE and its reduced versions on BLOG5K w.r.t. Micro-F1 and Macro-F1 scores. LAAE-minus-latent indicates removing latent ties from LAAE and LAAE-minus-adversarial indicates removing adversarial regularization from LAAE.}
  \label{fig:blogcatalog_ablation}
\end{figure}

\begin{figure}[htbp]
  \centering
  \includegraphics[width=1\textwidth,trim={50 0 50 0},clip]{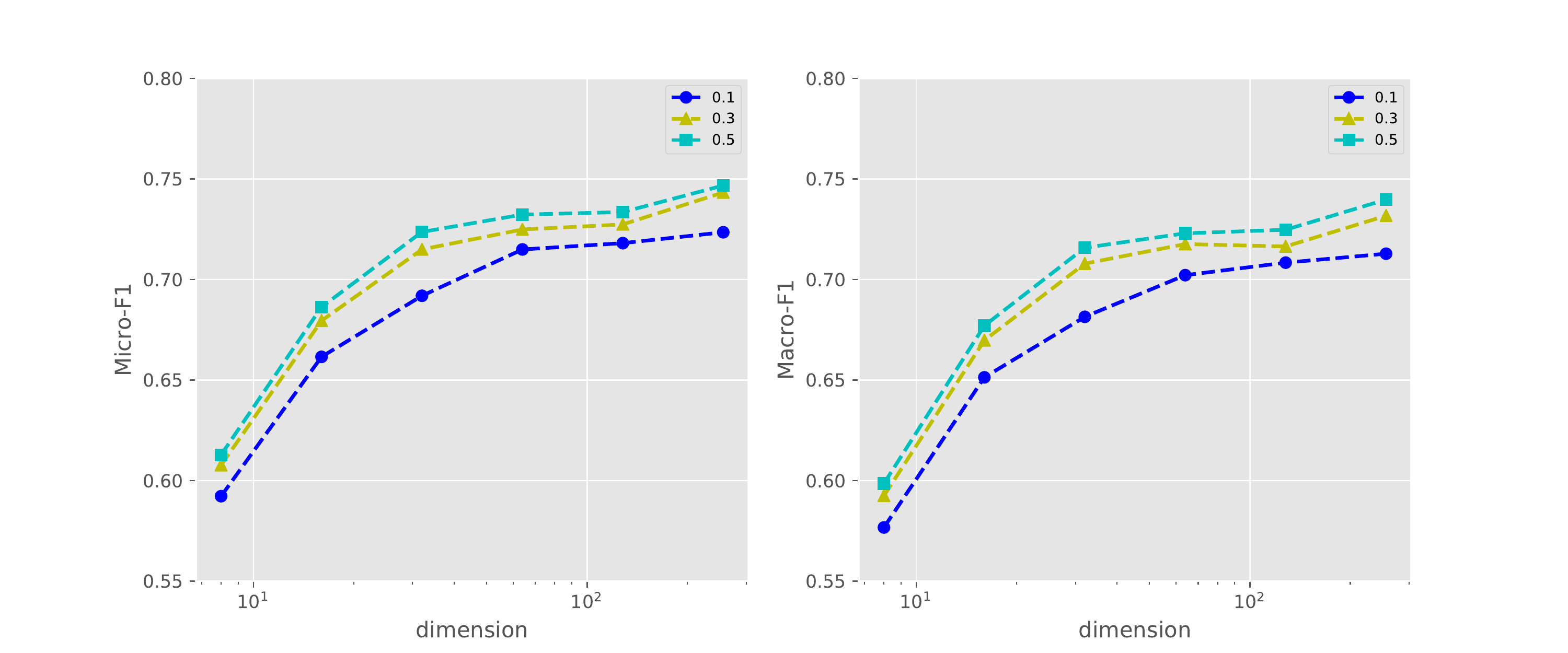}
  \caption{Performances in different dimensions ($d=8, 16, 32, 64, 128, 256$) on BLOG5K w.r.t. Micro-F1 and Macro-F1 scores.}
  \label{fig:blogcatalog_dim}
\end{figure}

To further understand the roles of different parts in our model, we conduct several experiments to analyze our model from different perspectives. In this subsection, we would first like to answer the following questions: (1) How does latent structure learning affect the performances. (2) What is the role of adversarial regularization in our model. To answer these two questions, we separately remove the latent inference part and adversarial part and record the performance obtained in BLOG5K. The results of ablation experiments are plotted in \reffig{fig:blogcatalog_ablation} accordingly. LAAE-minus-latent is the model that removes the latent ties from LAAE. The adjacent matrix $A$ without any latent information is directly fed into the adversarially regularized auto-encoders and the embeddings are learned by optimizing Eq. \eqref{loss}. LAAE-minus-adversarial represents the model that removes the adversarial regularization term from LAAE.

We observe that without latent ties, the performance drops dramatically w.r.t. both Micro-F1 and Macro-F1 scores. Besides, the results are easily affected by the shortage of training samples (i.e., training ratio is 0.1) when the latent ties are removed. The role of latent ties is to increase the information hidden behind the observed network, which can largely boost the quality of learned features and also the classification results. Since introducing latent ties relieves the sparsity issue in original networks, the performance is more stable when the training ratios are relatively small, which is of great meaning in real scenarios where the node labels are generally difficult to obtain. When removing the adversarial regularization, the results experience a slight decrease. The performance fluctuates when the training ratios changes. The role of adversarial regularization is to improve the feature robustness and model stability. The unfavorable noises brought during the process of inferring latent ties can be filtered by the adversarial training.

To analyze the parameters, we further study the influences of hidden dimensions in our model. The results of classification in BLOG5K with respect to different dimensions are plotted in \reffig{fig:blogcatalog_dim}. The performance ratchets up when the dimension increases and levels off when the dimension reaches 64.

\section{Conclusion}
We proposed in this paper a latent network embedding model based on adversarial auto-encoders, which served as a general framework for unsupervised representation learning in plain networks. First, the proposed model discovered latent ties from partial observations to capture the unobserved underlying structures. Second, our work aimed to distinguish the strong connections and weak connections of the latent graph by equipping the edges with different weights. The above steps provided a more comprehensive and precise way to describe the network structures. The experiments on link prediction and node classification showed that the introduction of latent information can boost the performance. Third, to cope with the possible noises in latent networks, we used an adversarial regularization term to increase the robustness of learned features. The ablation study showed that the adversarial term can improve the stability of learned features when varying the ratio of training samples.

The future direction of this paper first point to developing an end-to-end framework that can jointly optimize the structure learning phase and feature learning phase. It is also promising to apply our work to more complex graph types such as attribute graphs and heterogeneous networks. For example, consider a heterogeneous graph in recommendation systems where nodes have different types, inferring a latent graph not only provides more information about the similarities of users or items, but also the underlying preferences between users and items.

\bibliographystyle{ieeetr}
\bibliography{references}  

\end{document}